%% file: main.tex
\def\year{2024}\relax
\documentclass[letterpaper]{article} 
\usepackage{aaai24}  
\usepackage{times}  
\usepackage{helvet}  
\usepackage{courier}  
\usepackage[hyphens]{url}  
\usepackage{graphicx} 
\usepackage{natbib}  
\usepackage{caption} 
\usepackage{algorithm}
\usepackage{algorithmic}

\usepackage{newfloat}
\usepackage{listings}
\DeclareCaptionStyle{ruled}{labelfont=normalfont,labelsep=colon,strut=off} 
\floatstyle{ruled}
\newfloat{listing}{tb}{lst}{}
\floatname{listing}{Listing}
\usepackage{amsfonts} 
\usepackage{amsmath}
\usepackage{amssymb}
\usepackage{multirow} 
\usepackage{makecell}
\usepackage{subfigure}  
\usepackage[switch]{lineno}
\newcommand\ourmethod{{LogFormer}}
\usepackage{microtype}
\usepackage{soul}
\usepackage{amsthm}
\usepackage{booktabs}
\usepackage{subfig}
\usepackage{enumerate}
\usepackage{latexsym}
\usepackage{array}
\usepackage{comment}
\usepackage{epstopdf}
\usepackage{mathrsfs}
\usepackage{xcolor}

\usepackage{newfloat}
\usepackage{listings}

\floatstyle{ruled}
\newfloat{listing}{tb}{lst}{}
\floatname{listing}{Listing}
\title{\textsc{LogFormer:} A Pre-train and Tuning Pipeline for Log Anomaly Detection}
\author
{Hongcheng Guo \textsuperscript{1}, Jian Yang \textsuperscript{1}, Jiaheng Liu \textsuperscript{1}, Jiaqi Bai \textsuperscript{1}, Boyang Wang \textsuperscript{1}, Zhoujun Li \thanks{Corresponding Author} \textsuperscript{1}\\Tieqiao Zheng \textsuperscript{2}, Bo Zhang
\textsuperscript{2}, Junran peng \textsuperscript{3}, Qi Tian \textsuperscript{4}\\
\textsuperscript{1}Beihang University,
\textsuperscript{2}Cloudwise Research, 
\textsuperscript{3}Cheery.AI,
\textsuperscript{4}Huawei\\
{\tt\small \{hongchengguo,jiaya,liujiaheng,bjq,wangboyang,lizj\}@buaa.edu.cn\\ \{steven.zheng,bowen.zhang\}@cloudwise.com
}
}

\begin{document}
\maketitle

\begin{abstract}
Log anomaly detection is a key component in the field of artificial intelligence for IT operations (AIOps). Considering log data of variant domains, retraining the whole network for unknown domains is inefficient in real industrial scenarios. However, previous deep models merely focused on extracting the semantics of log sequences in the same domain, leading to poor generalization on multi-domain logs. To alleviate this issue, we propose a unified Trans\textbf{former}-based framework for \textbf{Log} anomaly detection (\ourmethod{}) to improve the generalization ability across different domains, where we establish a two-stage process including the pre-training and adapter-based tuning stage. Specifically, our model is first pre-trained on the source domain to obtain shared semantic knowledge of log data. Then, we transfer such knowledge to the target domain via shared parameters. Besides, the Log-Attention module is proposed to supplement the information ignored by the log-paring. The proposed method is evaluated on three public and one real-world datasets. Experimental results on multiple benchmarks demonstrate the effectiveness of our \ourmethod{}  with fewer trainable parameters and lower training costs.\footnote{https://github.com/HC-Guo/LogFormer.}
\end{abstract}

\input{Introduction}
\input{Related_work}

\input{Approach}

\input{Experiment}

\input{Conclusion}
\clearpage
\input{Acknowledgement}


\bibliography{ref}

\end{document}

%% file: Introduction.tex
\section{Introduction}

With the rapid development of large-scale IT systems, numerous companies have an increasing demand for high-quality cloud services. Anomaly detection \cite{breier2015anomaly} is critical to monitor data peculiarities for logs, which describe detailed system events at runtime and the intention of users in the large-scale services \cite{zhang2015rapid}. It is error-prone to detect anomalous logs from a local perspective. In this case, some automatic detection methods based on machine learning are proposed \cite{xu2009detecting}. Due to the development of IT services, the volume of log data has grown fast and traditional approaches are infeasible. Meanwhile, as log messages are half-structured and have their semantics, it is similar to natural language corpus. Therefore, many deep learning methods based on language models~\cite{hochreiter1997long,devlin2018bert} have been proposed on log anomaly detection task~\cite{zhang2016automated,du2017deeplog,zhang2019robust,meng2019loganomaly,loglg}. However, these models adopt parser~\cite{spell2016,2017drain} to gain templates in logs before detection, which leads to the loss of semantics in raw log data.

\begin{figure}[t]
    \centering {
\includegraphics[width=0.9\columnwidth]{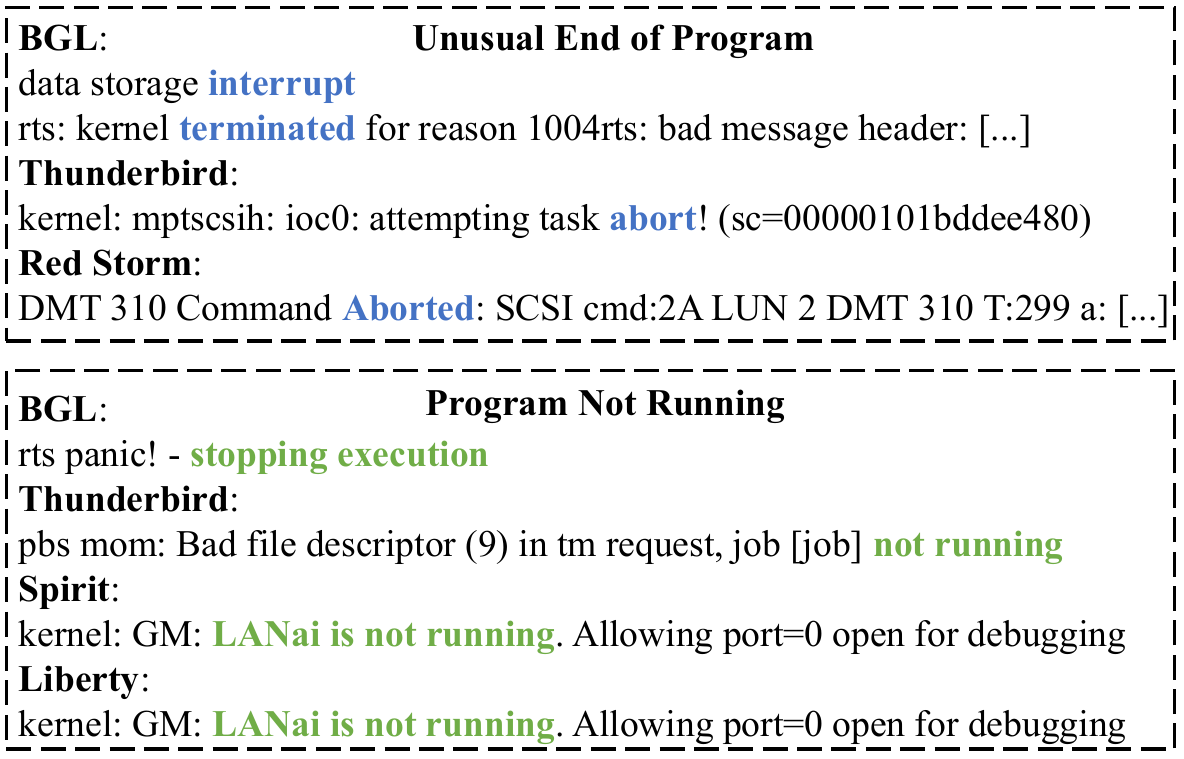}
    }

    \caption{The same anomaly from multiple domains. The top part denotes the ``Unusual End of Program'' anomaly from three domains including BGL, Thunderbird, and Red Storm while the bottom part is the ``Program Not Running'' from four domains including BGL, Thunderbird, Spirit, and Liberty.}
    \label{anomalyindomain}

\end{figure}

Despite being different in morphology and syntax, logs of multiple domains usually share similar semantic space. For example, in Fig.~\ref{anomalyindomain}, three sources (BGL, Thunderbird, Red Storm) have the same anomaly called \textbf{Unusual End of Program}. 
However, existing methods mostly focus on a single domain. When the components from a new domain are introduced, these methods lack the ability to accommodate such unseen logs. Besides, we need to consider the continuous iteration of log data when the system upgrades and it is costly to retrain different models for different datasets.

In this paper, we address the problems above via a two-stage solution called \ourmethod{}. \ourmethod{} is capable of preserving the shared semantic knowledge between different domains.
Specifically, to avoid information loss due to parsing, Log-Attention module is to supplement the information ignored by the log-paring for better performance. Then in the first stage, we create a model based on the Log-Attention, which is pre-trained on the source domain to obtain common semantics of log sequences. Second, \ourmethod{} uses a flexible component called Adapter to transfer knowledge from the source domain to the target domain. 

Generally, the contributions are as follows: (i) We propose \ourmethod{}, an end-to-end Pre-train and Tuning pipeline to automatically detect log anomalies,
which provides a new perspective via simple and effective pre-training and adapter-based tuning strategies for log anomaly detection. (ii)
Log-Attention module is proposed to avoid the loss of semantics caused by log parsing.
(iii) With only a few additional trainable parameters on the target domain,
the training costs are reduced a lot based on the effective parameter-sharing strategy in \ourmethod{}.
(iv) \ourmethod{} achieves state-of-the-art performance on three public benchmark datasets.

%% file: Related_work.tex
\section{Related Work}
\begin{figure}[ht]
    \centering {
        \includegraphics[width=0.9\columnwidth]{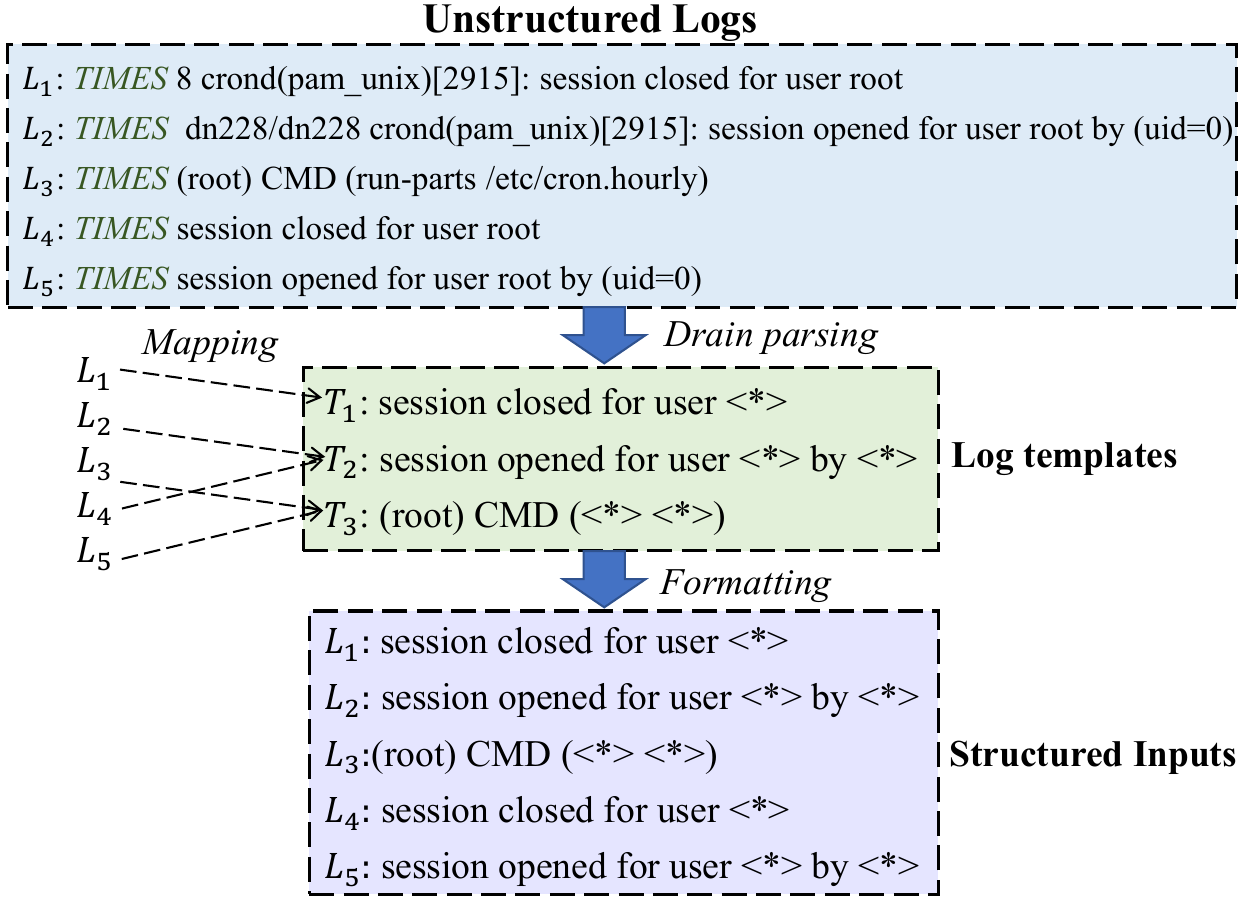}
    }

    \caption{Logs and Templates. The top part is unstructured logs, we adopt Drain algorithm to extract log templates,then we match each log with its template, which is the middle part. The bottom part is structured inputs.  }
 
    \label{preprocessing}
\end{figure}

\paragraph{Log Parsing}
Developers can create an unlimited number of variable names, abbreviations, or special technical terms,
which are out of the scale of ordinary English words. If we conduct word splitting, the endless unseen log tokens would explode the vocabulary, which is called the out-of-vocabulary (OOV) problem. To handle this issue, log parsing is used to convert unstructured logs into structured event templates by \textbf{keeping keywords}~\cite{AEL,Iplom,2017drain} and \textbf{removing extra parameters}, where the parameters usually denote special fields (e.g., /etc, /tmp), words (e.g., *, \_), serial numbers (e.g., 0x10001) and so on. In Fig.~\ref{preprocessing}, we use Drain~\cite{2017drain} to extract the templates, and each log  and the corresponding template are matched. Then, the log template sequence is fed into anomaly detection models.

\paragraph{Log Anomaly Detection}
Natural language processing technology~\cite{griprank,um4,lvpm3} is evolving rapidly, which has also reinvigorated this task. Supervised methods are based on classification. \cite{breier2015anomaly,hitanomaly,lu2018detecting}. LogRobust \cite{zhang2019robust} uses both normal and abnormal log data for training based on the Bi-LSTM architecture. Some semi-supervised methods \cite{xu2009detecting,semi-supervisedloganomaly} are proposed to alleviate such burden. DeepLog \cite{du2017deeplog} uses LSTM  to forecast the next log sequence with ranked probabilities. Besides, LogAnomaly \cite{meng2019loganomaly} uses  log  embeddings to capture the semantic information. PLElog \cite{PLELog} clusters the features of normal data and detects the anomalies by GRU. Although these methods obtain performance improvements on existing log datasets from a single source, they ignore the shared semantics between multiple sources and the value of the parameters removed by log parsing.

%% file: Approach.tex
\section{Approach}

\begin{figure*}[t]
    \centering {\includegraphics[width=0.8\textwidth]{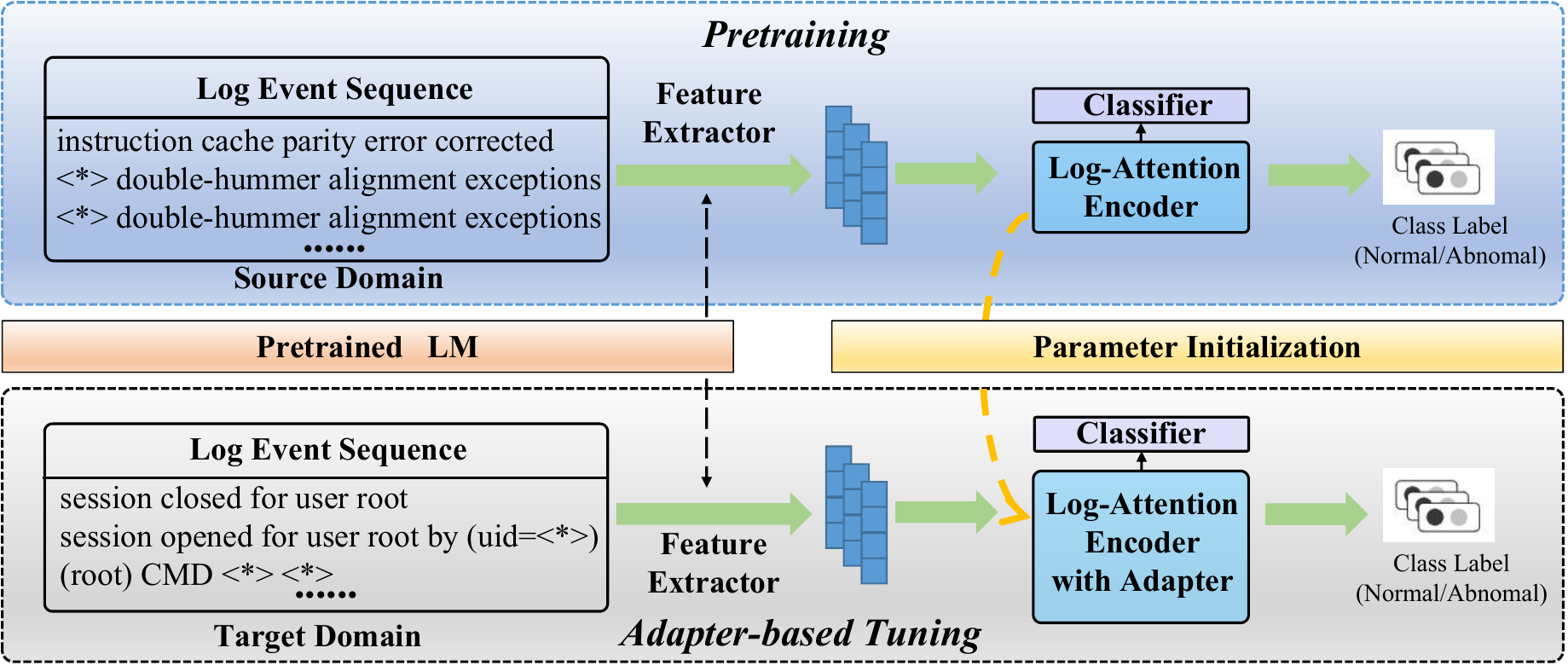}}

    \caption{Overview of architecture. Log sequences are first fed into the pre-trained language model to extract features. The Log-Attention encoder is trained on the source domain to acquire shared semantic information. Then, we initialize the encoder and only tune the parameters of the adapter on the target domain to transfer the knowledge.}
    \label{fig1}
\end{figure*}

In this section, we describe the general framework of \ourmethod{}. In Fig.~\ref{fig1}, \ourmethod{} contains two stages: pre-training and adapter-based tuning. In the following, we present the definition of the problem, and the components of \ourmethod{}. Finally, we introduce the processes of the two stages.

\subsection{Problem Definition}

Log anomaly detection problem is defined as a binary classification task. The model is supposed to determine whether the input log is abnormal or normal. For the source domain, after preprocessing each raw logs, we generate the vector representations of $K_{src}$ log sequences, which are denoted as $S^{src}=\{ S_{k}\}_{k=1}^{K_{src}}$. Then, $S^{src}_{i}=\{V_{t}^{src}\}_{t=1}^{T_i^{src}}$ denotes the $i$-th log sequence, where $T_i^{src}$ is the length of the $i$-th log sequence and $V_{t}^{src}$ denotes the $t$-th log sentence in $S^{src}_{i}$.
For the target domain, $S^{tgt}= \{S^{tgt}_{k}\}_{k=1}^{K_{tgt}}$ denotes the representations of $K_{tgt}$ log sequences. $S^{tgt}_{j}=\{V_{t}^{tgt}\}_{t=1}^{T_j^{tgt}}$ denotes the $j$-th log sequence, where $T_j^{tgt}$ is the length of the $j$-th log sequence. Therefore, the training procedure is as follows. We first pre-train the model on the dataset from source domain as follows:
\begin{equation}
    f_{p}(y_{i}|S^{src}_{i};\Theta)), 
    \label{eq:1}
\end{equation}where $f_{p}$ represents the pre-training stage and $\Theta$ is the model parameters in pre-training stage. Then, the model is transferred to the target domain:
\begin{equation}
    f_{a}(y_{j}|S^{tgt}_{j};\Theta_{f},\theta_{a}). 
    \label{eq:2}
\end{equation}where $f_{a}$ represents the adapter-based tuning stage. $\Theta_{f}$ denotes the parameter of the encoder transferred from the pre-training stage, which is frozen in adapter-based tuning stage. $\theta_{a}$ is the parameter of the adapter. $y$ is the ground-truth label.

\subsection{Feature Extractor}
%


The feature extractor converts session sequences (template sequence) to vectors with the same dimension $d$. Here we use the pre-trained sentence-bert~\cite{reimers2019sentence} model to get the representation of the template sequence in Fig.~\ref{preprocessing}. Suppose each session has $l$ fixed length, then the embedding of the input $X$ after the feature extractor:

\begin{equation}
    X_{E} = \mathbf{FE}(X).
\end{equation} Where \textbf{FE} represents the encoder of sentence-bert. we can obtain $X_{E} \epsilon R_{}^{l \times d}$ for each session. 

\subsection{Log-Attention Module}
\begin{figure}[ht]
    \centering {
        \includegraphics[width=0.9\columnwidth]{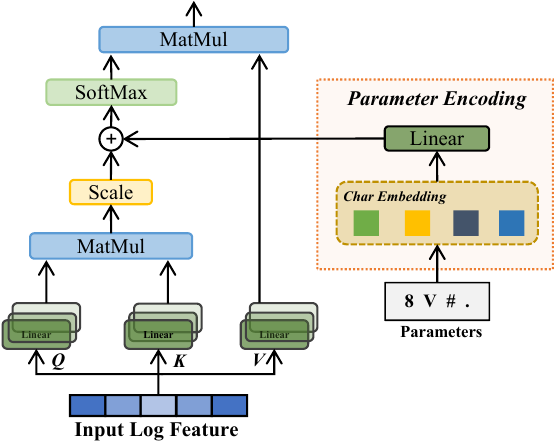}
    }

    \caption{Log-Attention. The left part is the multi-head attention, and the right part is the parameter encoding.}
    \label{logattention}

\end{figure}

Although parsing solves the  out-of-vocabulary (OOV) problem, the information of the parameters is discarded. To aggregate parameters and keywords information, we have adjusted the structure of the original transformer encoder. Log-Attention module is proposed in Fig.~\ref{logattention}. Specifically, after parsing, we gain the $P$ parameters for each log sequence. For each character $P_{i}$ in $P$, we adopt the feature extractor to obtain character-level embedding $P_{i}^{E}$. Then we use the Linear layer to encode the whole $P^{E}$ as follows:
\begin{equation}
    \phi_{p} = \mathbf{LINEAR}(P^{E}).
\end{equation}
where $\phi_{p}$ denotes the output of parameter encoding. Then, we assign each output a learnable scalar, which will serve as a bias term in self-attention.
The intuition of Log-Attention is also inspired by position encoding,
which is mapped as bias in attention and provides additional position information. 
The Log-Attention is computed as follows:


\begin{equation}\small
    LogAttention=\mathbf{Softmax}(\frac{QK_{}^{T}}{\sqrt{d/h} } + \phi_{p})V.
\end{equation}
where $h$ is the number of the heads, $d$ denotes the dimension of the input, and $Q,K,V$ represent queries, keys, and values, respectively.

\subsection{Encoder with Adapter}

\begin{figure}[ht]
    \centering {
        \includegraphics[width=0.85\columnwidth]{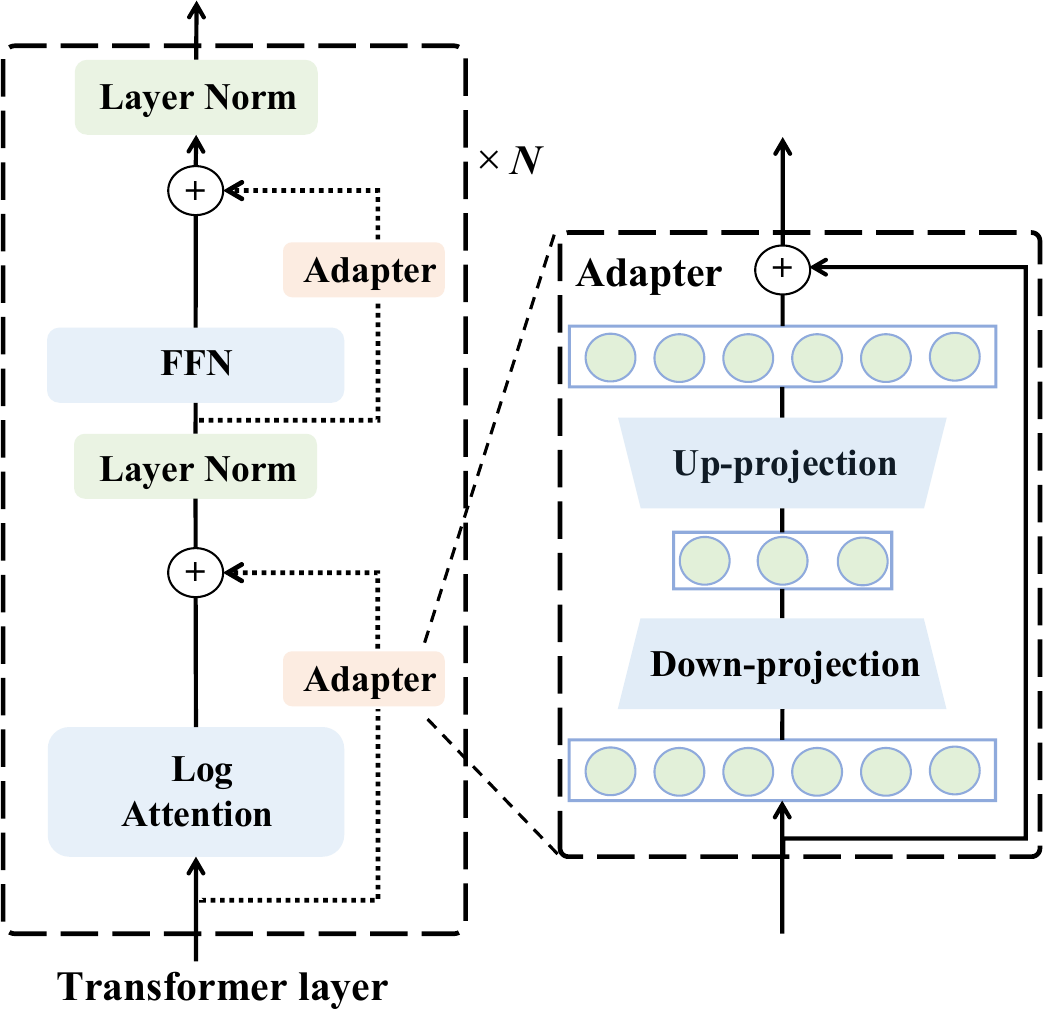}
    }

    \caption{Encoder with Adapters. Where $N$ is the number of encoder layers. The left part describes the log-attention encoder inserted by parallel adapters, and the right part is the structure of an adapter, which is composed of the down- and up-projection layers.}
    \label{adapter}

\end{figure}

The order of a log sequence conveys information about the program execution sequence. Wrong execution order is also considered abnormal. Thus, constant positional embedding is also used. The component after the attention layer and feedforward layer is the original serial adapter. We design our log adapter with a parallel structure in Fig.~\ref{adapter}, which is inserted parallel to the Log-Attention layer and feedforward layer. This design allows adapter to use input information better with original complete encoders. During adapter-based tuning, only a few parameters of the adapters are updated on the target domain. More specifically, we use down- and up-scale neural networks as the adapter. Two projection layers first map the hidden vector from dimension $d$ to dimension $m$ and then map it back to $d$. The adapter also has a skip-connection operation internally. The output vector $h'$ is calculated:
\begin{equation}
    h'= W_{up} tanh(W_{down}h)+h.
\end{equation}
where $h \in \mathbb{R}^d$ represents a given hidden vector.  $W_{down} \in \mathbb{R}^{m\times d}$ and $W_{up} \in \mathbb{R}^{d\times m}$ is the down-projection and the up-projection matrix respectively, by setting $m<<d$, we limit the number of parameters added per adapter, which is the core to reduce trainable parameters while retaining semantic information.

\subsection{Pre-training}
Inspired by pre-trained models~\cite{devlin2018bert,reimers2019sentence,owl}, we can acquire the common representation for log anomaly with the stacked log-attention encoders. In this stage, the pre-trained model learns the commonalities among different anomalies. 
Specifically, the objective of pre-training is a supervised binary classification task without using log adapter. Then, the parameters of the log-attention encoders are used as the initialization of the next stage.

\subsection{Adapter-based Tuning}
\label{adapter_based_tuning}
Adapter-based tuning leverages the knowledge obtained from the pre-training  with lightweight adapters~\cite{houlsby2019parameter}. Specifically, in the second stage, we plug adapters into the encoders of the pre-trained model, where only the parameters of the adapters are updated during target domain adaption. Parameters of the Log-Attention and the feedforward layers in the pre-trained model are frozen. Unlike fine-tuning, \ourmethod{} provides a plug-in mechanism to reuse the pre-trained model with only a few additional trainable parameters.



\subsection{Training Strategy}
The classifier is simply implemented by one linear layer. We both take BCE loss for two stages. Thus, the loss of the pre-training stage  is as follows:
\begin{equation}
       \mathcal{L}_{p} = -\mathbb{E}_{x,y \in D^{src}_{x,y}} [\log P(y|x;\Theta) ], 
\end{equation}where $\mathcal{L}_{p}$ represents the loss in the pre-training stage. $\Theta$ is the parameter of the whole model in the pre-training stage. $x$ and $y$ are the input data and label respectively, $D^{src}_{x,y}$ represents the data coming from the source domain. 
Then, we define the objective loss in the adapter-based tuning stage: 
\begin{equation}
       \mathcal{L}_{a} = -\mathbb{E}_{x,y \in D^{tgt}_{x,y}} [\log P(y|x;\Theta_{f},\theta_{a}) ]. 
\end{equation}where $\mathcal{L}_{a}$ is the loss function in the adapter-based tuning stage. $\Theta_{f}$ is the parameter of the encoder module trained in the pre-training stage, which is frozen in the adapter-based tuning stage. $\theta_{a}$ is the parameter of the adapter. $D^{tgt}_{x,y}$ represents the data coming from the target domain.

%% file: Experiment.tex
\section{Experiments}
In this section, we compare our method with existing methods on multiple benchmark datasets.

\paragraph{Datasets} 
We conduct experiments on three datasets from LogHub~\cite{he2020loghub} \footnote{https://github.com/logpai/loghub}.
HDFS~\cite{xu2009detecting} dataset is generated and collected from the Amazon EC2 platform through running Hadoop-based map-reduce jobs. Thunderbird and BGL datasets~\cite{oliner2007supercomputers} contain logs collected from a two-supercomputer system at Sandia National Labs (SNL) in Albuquerque. The log contains alert and non-alert messages identified by alert category tags. Following~\cite{yao2020study,meng2019loganomaly}, 10M/11M/5M continuous log messages from Thunderbird/HDFS/BGL are used.
\begin{table}[h]
    \begin{center}
        \resizebox{1\columnwidth}{!}{
        \begin{tabular}{lccccc}
            \toprule
            Dataset   & Category   &   \#Messages     & \#Anomaly & \#Templates & \#Error Types  
            \\
            \midrule
            HDFS   &  Distributed   &    11M  &  17K & 49  & 53
            \\
            BGL    &    Supercomputer     & 5M       & 40K   & \textbf{1423}   &\textbf{143} 
            \\
            Thunderbird   & Supercomputer  &  10M    & 123K  & 1092 & 95
            \\
            \bottomrule
        \end{tabular}}
        \caption{A summary of the datasets used in this work. Messages are the raw log strings. Log sequences are extracted by ID or sliding window method.}
        \label{table1}
    \end{center}
\end{table}

\paragraph{Preprocessing}

We extract log sequences by \textbf{block IDs} in HDFS. For BGL and Thunderbird, we utilize the \textbf{sliding window} (size of 20) without overlap to generate log sequences. We adopt Drain~\cite{2017drain} with specifically designed regex to do log parsing. For each dataset, considering that logs evolve over time, we select the first $80\%$ (according to the timestamp of logs) log sequences for training and the rest $20\%$ for testing, which is consistent with the prior work~\cite{PLELog,du2017deeplog}.

\paragraph{Implementation Details}

In experiments, we use different numbers of transformer encoder layers in $\{1,2,4\}$. The number of attention heads is 8, and the size of the feedforward network that takes the output of the multi-head self-attention mechanism is 3072. 
We use Adam as the optimization algorithm whose learning rate is scheduled by OneCycleLR, with $\beta_1 = 0.9$, $\beta_2 = 0.99$, and $\varepsilon = 10^{-8}$. All runs are trained on 2 NVIDIA A100(40G) with a batch size of 64. For each dataset, we tune the maximum learning of the OneCycleLR scheduler in $\{1e-5,5e-5,1e-6\}$.



\paragraph{Baselines and Evaluation}


 Table~\ref{table1} shows five public baselines including Support Vector Machine(SVM), Deeplog \cite{du2017deeplog}, LogAnomaly \cite{meng2019loganomaly}, LogRobust \cite{zhang2019robust}, and PLELog \cite{PLELog} \footnote{https://github.com/YangLin-George/PLELog} and two variants of \ourmethod{}. We also compare with the popular ChatGPT~\cite{chatgpt}. Here, \ourmethod{}$_{S}$  is trained from scratch without two stages. \ourmethod{}$_{P}$ is trained only with the pre-training stage, which means we directly tune the whole parameters from the pre-trained model. For a fair comparison, these baselines are trained on the union of source and target domain, as \ourmethod{} utilize the knowledge of source domain (BGL) and target domain (HDFS/Thunderbird).
For evaluation, we use Precision ($\frac{TP}{TP+FP}$), Recall ($\frac{TP}{TP+FN}$) and $F_1$ score ($\frac{2*Precision*Recall}{Precision+Recall}$).

\paragraph{Main Results}

\begin{table}[t]
    \centering
    \begin{center}
    \resizebox{0.95\columnwidth}{!}{
    \begin{tabular}{l c c c c}
        \toprule
        Dataset & Method & Precision & Recall & $F_1$ Score\\
        
        \midrule

        &SVM  &0.31  &0.65  &0.41  \\
        &DeepLog  & 0.83  & 0.87  & 0.85 \\
        &LogAnomaly & 0.86  & 0.89 & 0.87 \\
        &PLELog & 0.88 & 0.93 & 0.90\\
        HDFS&LogRobust  & 0.88  & 0.95  & 0.91 \\
        &ChatGPT  & 0.74  & 0.82  & 0.78 \\
        &\ourmethod${}_{S}$  & 0.95  & 0.96  & 0.95 \\
        &\ourmethod{}$_{P}$  & 0.96  & 0.97  & 0.96 \\
        &\ourmethod{}  & \textbf{0.97}  & \textbf{0.98} & \textbf{0.98} \\
            
        \midrule
        
        &SVM &0.22 &0.56 &0.32\\
        &DeepLog&0.14  &0.81  &0.24\\
        &LogAnomaly & 0.19 &0.78 &0.31\\
        BGL&PLELog & 0.92 & 0.96 & 0.94\\
        &LogRobust& 0.92 & 0.96 & 0.94\\
        &ChatGPT  & 0.77  & 0.71  & 0.74 \\
        &\ourmethod${}_{S}$ & \textbf{0.96} & \textbf{0.97} & \textbf{0.97}\\

        \midrule
        
        &SVM & 0.34 & 0.91 &0.46\\
        &DeepLog & 0.48  & 0.89 & 0.62\\
        &LogAnomaly & 0.51 & 0.97 &0.67\\
        &PLELog &0.85  &0.94  &0.89 \\
        Thunderbird&LogRobust & 0.89 & 0.96 &0.92\\
        &ChatGPT  & 0.84  & 0.79  & 0.81 \\
        &\ourmethod${}_{S}$ & 0.94 & 0.98 & 0.96\\
        &\ourmethod${}_{P}$ & 0.97 & 0.99 & 0.98\\
        &\ourmethod{} & \textbf{0.99} & \textbf{0.99} & \textbf{0.99}\\
  
        \bottomrule
    \end{tabular}}

    \caption{Results on Thunderbird, BGL and HDFS. \ourmethod{}$_{S}$ represents the model trained from scratch, \ourmethod${}_{P}$ represents the model trained with pre-training but tuning without adapters.}
    \label{table2}
    \end{center}

\end{table}

In Table~\ref{table2}, baselines are trained on the union of source and target domain data for a fair comparison. In our setting, BGL dataset is chosen as the source domain for the pre-training, HDFS and Thunderbird are chosen as the target domain. \ourmethod{} achieves the highest $F_1$ score on all three settings. Specifically,
results show that most baseline methods perform badly when BGL data is used for training. It is reasonable for the diverse types of error and complex structure of logs in BGL. This also confirms that these baselines have poor generalizability and cannot handle multi-source logs together. When only BGL data is used for training, LogRobust and PLElog achieve a comparable $F_1$ score with \ourmethod{}$_{s}$, this means our backbone model with Log-Attention module is strong enough without pre-training and adapter-tuning on the single source. 

\paragraph{Time Consumption}
Table~\ref{time consumption} shows the 
training and testing time of \ourmethod{} on HDFS, BGL, and Thunderbird, respectively. \ourmethod{} gains the lowest training and testing time consumption compared with these state-of-the-art methods.


\begin{table}[]
\centering
\resizebox{0.95\columnwidth}{!}{
\begin{tabular}{c|cc|cc|cc}
\toprule
\multirow{2}{*}{Method} & \multicolumn{2}{c|}{HDFS} & \multicolumn{2}{c|}{BGL} & \multicolumn{2}{c}{Thunderbird} \\
                        & Training    & Testing     & Trainging    & Testing   & Training        & Testing       \\ \midrule
                        DeepLog & 50m       & 10m     &23m        & 6m         &  59m               & 12m  \\ 
                        LogAnomaly & 1h 48m        & 22m    &  1h 10m       &   20m        &1h 43m             & 30m \\ 
                        PLElog & 42m        &  35s    & 20m      & 14s                      & 36m        & 31s  \\ 
                        LogRobust & 1h 01m     &  17m    & 40m      & 4m                       & 58m        & 12m  \\ 
                        \ourmethod{} & \textbf{29m}   & \textbf{20s}   &    \textbf{17m}     & \textbf{11s}               &  \textbf{31m}               & \textbf{16s}   \\ 
                        \bottomrule
\end{tabular}}

\caption{Time consumption of different approaches. The lowest results are highlighted.}

\label{time consumption}
\end{table}

\section{Ablation Study} \label{analysis}

\begin{figure}[ht]
    \centering {
        \includegraphics[width=0.95\linewidth]{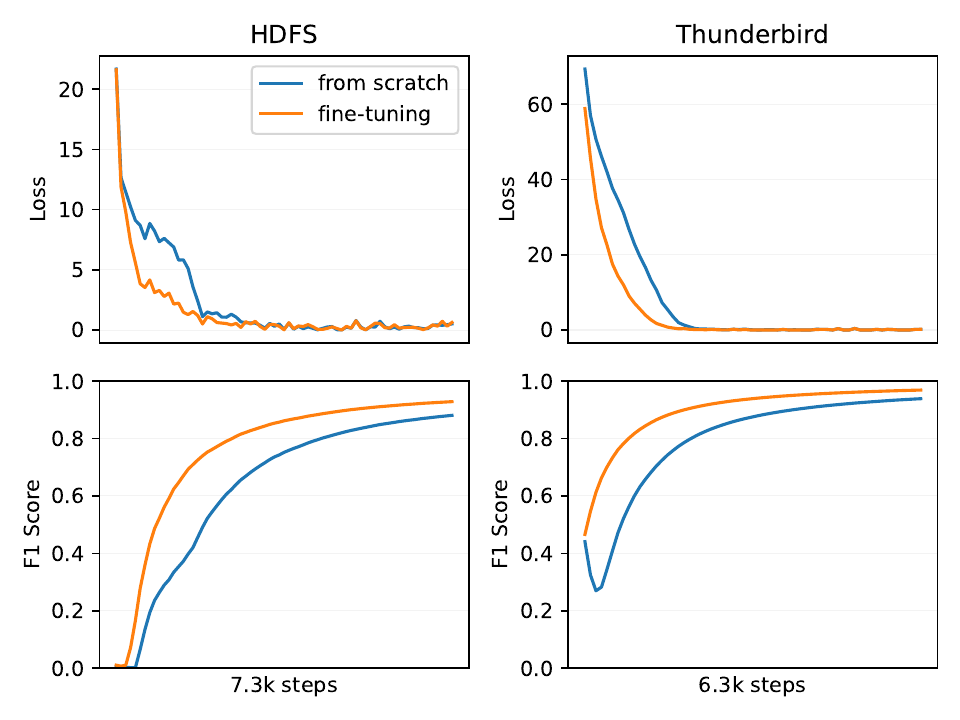}
    }

    \caption{Loss and $F_1$ score on the test set. We compare two ways of training including training from scratch and fine-tuning from the pre-trained model on BGL.}

    \label{fig:transfer}
\end{figure}

\paragraph{Effect of pre-training} \label{effectlogmodel}
To demonstrate the effectiveness of pre-training, we compare the performance of \ourmethod{}$_{s}$ and \ourmethod{}$_{p}$. We choose BGL as the source domain as its variety in log templates. We compare two strategies in terms of loss and $F_1$ score. Fig~\ref{fig:transfer} shows the loss and $F_1$ score curves in the training process (i.e., training steps). Results show that fine-tuning converges faster than training from scratch, which shows the learned knowledge from the source domain is valuable. Besides, the method with fine-tuning achieves higher performance in the initial stage and the loss curve is more stable, which also shows the power of pre-training.
For the $F_1$, we observe that fine-tuning requires fewer training steps to gain the best results, which is noteworthy for reducing costs in industrial scenes. To sum up, the pre-training stage is valuable and allows the model to converge quickly with better results.

\paragraph{Effect of Adapter-based Tuning}

Although we have shown that pre-training could accelerate convergence without decreasing performance, fine-tuning is expensive and important. Thus, we adopt adapter-based tuning to acquire a compact model for log anomaly detection by adding a few additional trainable parameters. To show the effect of  adapter-based tuning,
we compare the performance of \ourmethod{}$_{p}$ and \ourmethod{} on the HDFS and Thunderbird datasets as shown in Table~\ref{tab:transfer}. We have the following observations. First, \ourmethod{} generates a little higher $F_1$ score (1$\%$ on average) than directly fine-tuning the pre-trained model on two datasets. Second, Adapter-based tuning adopts $3.5\%-5.5\%$ of the trainable parameters compared to direct fine-tuning. Third, more encoder layers for fine-tuning do not generate better results. In contrast, adapter-based tuning performs more robustly with more encoder layers.

\begin{table}
\centering
\resizebox{0.95\columnwidth}{!}{
\begin{tabular}{lrrrr}
\toprule
Method          & \#Layers & \#Parameters & HDFS & Thunderbird \\
\midrule

       & 1  &   7.2M   & 0.945  &  0.969    \\
Tuning                & 2  &   14.3M    & 0.962  &  0.981    \\
                & 4  &   28.5M    & 0.961  &  0.980    \\
                \midrule
             & 1  &   \textbf{0.4M}   &  0.957  &  0.972  \\
Adapter tuning               & 2  &   \textbf{0.6M}    &  0.974  &  0.987    \\
                & 4  &   \textbf{1M}    &  \textbf{0.981}  &  \textbf{0.998}    \\
  
\bottomrule
\end{tabular}}
\caption{Results between fine-tuning and adapter-based tuning. {\#Layers} is the number of encoder layers. {\#Parameters} is the number of trainable parameters.}

\label{tab:transfer}
\end{table}


\paragraph{Effect of Log-Attention}

We compare the results of the original self-attention with our Log-Attention. To avoid the interference of other factors, we use \ourmethod${}_{S}$ in Table.~\ref{attention comparison}. Results show that our Log-Attention achieves 3.6\% higher points on average than the self-attention on three datasets, which shows the effect of Log-Attention module. Meanwhile, it shows that variables (removed by log parsing) also provide valuable information for the anomaly detection. 

\begin{table}[ht]
\centering
\resizebox{0.85\columnwidth}{!}{
\begin{tabular}{cccc}
\toprule
           Attention & HDFS & BGL & Thunderbird \\
\midrule

   Self-Attention        & 0.911  &  0.943   & 0.935 \\
                 
                \midrule
            
  Log-Attention              & \textbf{0.952}  &  \textbf{0.974}   &   \textbf{0.962}     \\
            
\bottomrule
\end{tabular}}
\caption{$F_1$ scores between self-attention and Log-Attention. Experiments are based on 4 encoder layers.}
\label{attention comparison}

\end{table}

\paragraph{Effect of Variants of Adapters}

We compare the $F_1$ scores of the variants of adapters including 
LoRA~\cite{LoRA} and Parallel-Adapter~\cite{serial_parallel} in Table~\ref{adaper comparison}. Results show that in our task, all three types of adapter gains great performance on three datasets, demonstrating the effectiveness of adapter-based tuning stage.

\begin{table}
\centering
\resizebox{0.85\columnwidth}{!}{
\begin{tabular}{cccc}
\toprule
Adapter            & HDFS & BGL & Thunderbird \\
\midrule

Serial Adapter               & 0.982  &  0.971   & 0.992 \\
                 
                \midrule
            
Parallel  Adapter            & 0.980  &  0.969   &  0.988      \\
\midrule

LoRA              & 0.981  &  0.972   &   0.993     \\

\bottomrule
\end{tabular}}
\caption{$F_1$ scores between serial adapter and ours. Experiments are based on \emph{4} layers of log-attention encoder.}
\label{adaper comparison}
\end{table}

\paragraph{Effect on Low-resource Setting}

\begin{figure}[ht]
    \centering {
        \includegraphics[width=0.9\linewidth]{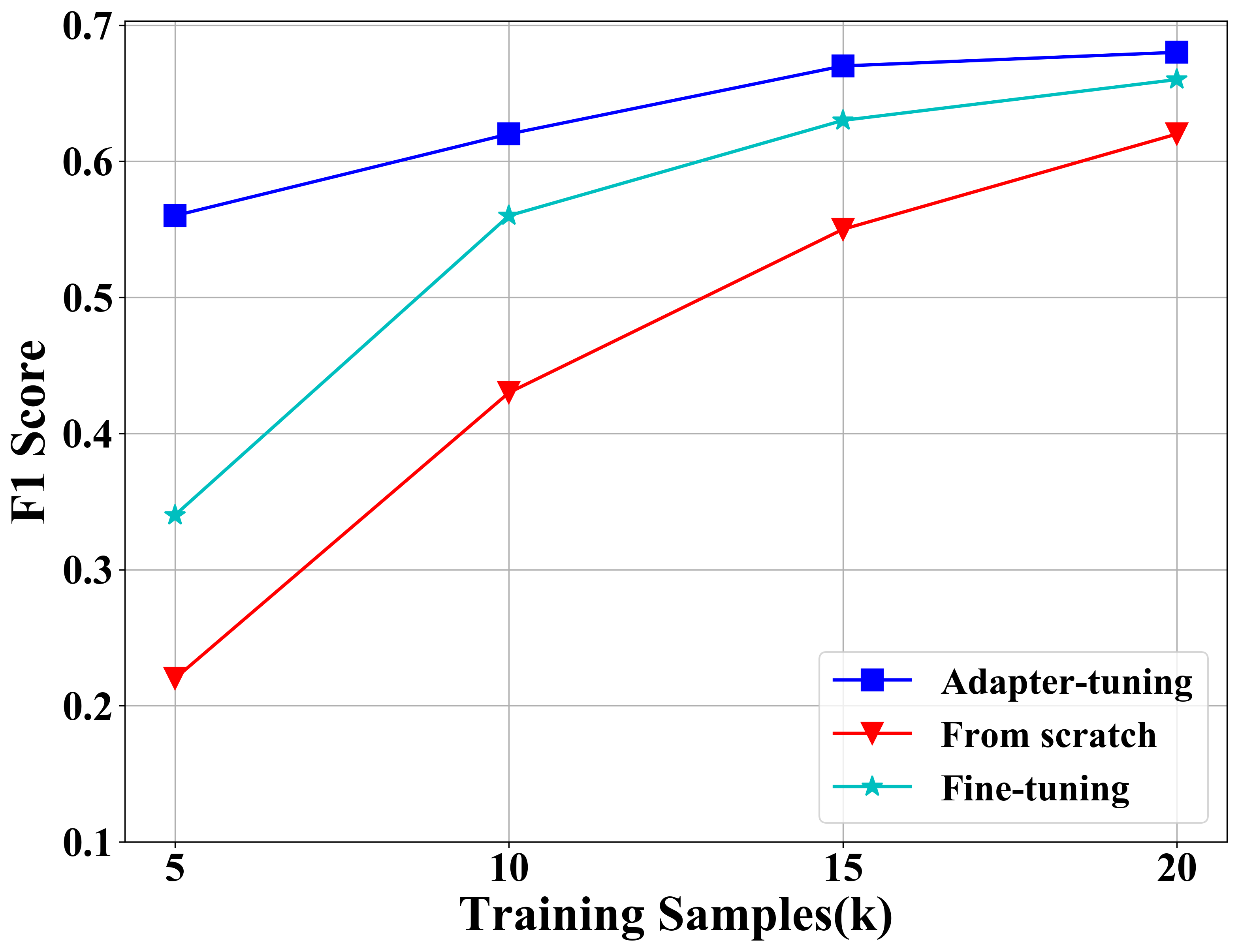}
    }
    \caption{Test performance on Thunderbird w.r.t.the number of training examples. 5k, 10k, 15k, 20k corresponding to the first $1.25\%$, $2.5\%$, $3.75\%$, $5\%$ training data respectively. We show $F_1$ scores for all methods.}
    \label{fig:low}
\end{figure}

To verify the power of \ourmethod{} under the low-resource setting, we consider the task with fewer than 20k training examples as the low-resource setting. The ablation study is conducted on the Thunderbird and models are sufficiently trained for 30 epochs. In Fig.~\ref{fig:low}, we compare the $F_1$ scores with different numbers of training samples ranging from 5k-20k. We find that 1) Adapter-based tuning consistently outperforms training and fine-tuning, especially when the training size is small. For example, we gain 34$\%$ improvements compared with training from scratch with only 5k data. 2) With the number of training samples increasing, the gap between the $F_1$ scores of all methods will become smaller. 3)  \ourmethod{} is robust, with a similar standard deviation across different training sizes. 
To summarize, \ourmethod{} provides acceptable results in the low-resource setting, which is highly parameter-efficient for log analysis. 

\begin{figure}[ht]
    \centering {
        \includegraphics[width=0.95\linewidth]{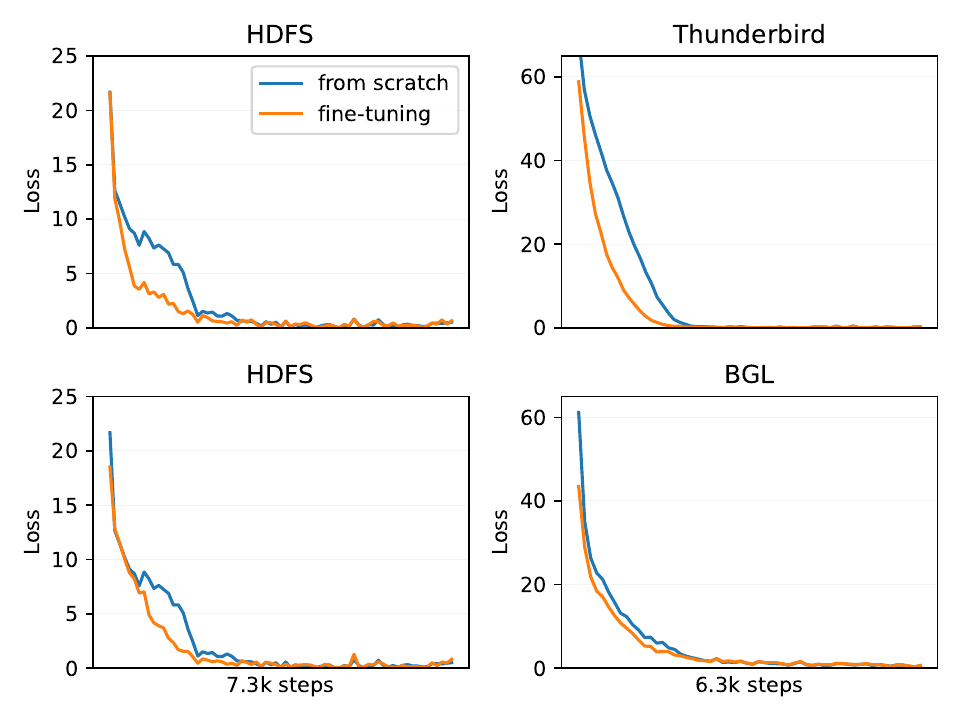}
    }
    \caption{Loss on the test set w.r.t training steps. The upper/bottom results are based on models pre-trained on BGL/Thunderbird. All results are based on using one layer encoder.}
    \label{fig:pre-trainedmodel}
\end{figure}

\paragraph{Effect of Source Domain} 
\ourmethod{} aims at transferring knowledge between domains to help detect log anomalies. Thus it is vital to choose the correct source domain for the pre-training stage. A suitable domain needs to meet two conditions: 1) Variety in templates and types of error. 2) For different and similar domains, it has the great power to migrate semantic knowledge.
HDFS has fewer templates and types of error compared with BGL and Thunderbird. Thus we do not utilize HDFS as the source domain. Specifically, we compare the results by choosing BGL and Thunderbird as the source domain respectively. In terms of the $F_1$ score, both of them gain high results on target domains. Thus we turn our attention to loss curves, Fig.~\ref{fig:pre-trainedmodel} shows the loss curves on the target domains. Comparing two pre-trained models,
on the HDFS dataset, the model pre-trained on BGL brings faster convergence. Besides, the model pre-trained on BGL brings faster convergence for Thunderbird than Thunderbird brings to BGL. Overall, BGL is the most suitable source domain for transferring semantics across domains.

\paragraph{Effect of Large Language Models}
To fully explore the ability of large language models to detect anomalous logs, we specially design a two-hop log Chain-of-Thought~\cite{chain-of-thought} (LCoT) approach for this task rather than directly generate the decision. Specifically, in Fig.~\ref{two_hop_Cot}, we first guide the model to generate the templates of the log sequences, than we concatenate the raw logs and templates as the prompts, and then let the model determine if there are any anomalies that arise. A comparison of the results using Chain-of-Thought and those generated directly is in Table.~\ref{COT_comparison}. We can see that large language models like ChatGPT can not perform well on log anomaly detection task. To explore why the large language model performs badly, we check the results and find that in most bad cases, the variables play an important role. Large language model is more concerned with sensitive words like "Error" and "failed". Such words contain obvious abnormalities, which can be recognized well. The findings further illustrate the importance of our log-attention, and in the future, we can inject external rules to make the large language models more sensitive to the variables.

\begin{figure}[t]\small
\renewcommand\arraystretch{1.2}
\centering
\begin{tabular}{rp{165pt}}
    \hline
    \hline
    \multicolumn{2}{c}{\textbf{Log Context}}  \\
    \hline
    
    \textcolor[rgb]{0.7,0,0}{\emph{log sequences:}} 
    & 2023-08-02 10:30:00 DEBUG: Checking server availability. \\
    & 2023-08-02 10:30:15 ERROR: NetworkException - Unable to establish connection to server. \\
    & 2023-08-02 10:31:00 INFO: Retrying connection to server.\\

    
    
    \hline
    \multicolumn{2}{c}{\textbf{Hop 1: Log Parsing}}  \\
    \hline
    
    \emph{One-hop Prompt:} & 
    Extract the templates of \textcolor[rgb]{0.7,0,0}{\emph{log sequences}} while replacing the \textcolor[rgb]{0,0.7,0}{\emph{variables}}with $<*>$\\

    \textcolor[rgb]{0,0,0.7}{\emph{Templates:}} & 1. 
     $<*>$ ERROR: NetworkException - $<*>$ to establish connection to server.\\
        & 2. $<*>$ DEBUG: Checking server availability. \\
    & 3. $<*>$ INFO: Retrying connection to server.\\

    \hline
    \multicolumn{2}{c}{\textbf{Hop 2: Anomaly Detection}}  \\
    \hline
    
    \emph{Two-hop Prompt:} & 
    According to the \textcolor[rgb]{0.7,0,0}{\emph{log sequences}}, \textcolor[rgb]{0,0,0.7}{\emph{Templates:}}, the relationship between \textcolor[rgb]{0,0,0.7}{\emph{Templates:}} and \textcolor[rgb]{0,0.7,0}{\emph{variables}}, determine if there are any exceptions in templates and variables, and directly give the answer: Yes or No.\\

    \emph{Answer:} & Yes or No. \\
    
    \hline
    \hline
\end{tabular}
\caption{An example of two hop log Chain-of-Thought process. We first extract the templates of the logs, then we let the ChatGPT find the anomalies according to the logs and templates.}
\label{two_hop_Cot}
\end{figure}

\begin{table}[ht]
\centering
\resizebox{0.95\columnwidth}{!}{
\begin{tabular}{cccc}
\toprule
           Method & HDFS & BGL & Thunderbird \\
\midrule

   ChatGPT w/o LCoT        & 0.78  &  0.74   & 0.81 \\
                 
                \midrule
    ChatGPT w/ LCoT        & 0.85  &  0.83   & 0.88 \\
                 
                \midrule
            
  \ourmethod{}             & \textbf{0.98}  &  \textbf{0.97}   &   \textbf{0.99}     \\
            
\bottomrule
\end{tabular}}
\caption{$F_1$ between ChatGPT with LCoT  and without LCoT.}
\label{COT_comparison}

\end{table}

\section{Practical Evaluation}

\ourmethod{} has been successfully applied to a cloud service company. To test the generalization of \ourmethod{}, we conduct experiments on a real-world distributed dataset called GAIA~\footnote{https://github.com/CloudWise-OpenSource/GAIA-DataSet}. As the online system serves hundreds of corporations, the generated logs are complex. Thus, it is difficult to detect anomalies on such multi-domain and continuously evolved data. Here, we take 8,200,000 log messages for the experiment (80$\%$ for training, 20$\%$ for testing), which consists of 31,279 anomalous messages. 
\begin{table}[t]
    \centering
    \begin{center}
    \resizebox{1\columnwidth}{!}{
    \begin{tabular}{l c c c c}
        \toprule
        Dataset & Method & Precision & Recall & $F_1$ Score\\
        \midrule
 
        &SVM  &0.21  &0.54  &0.30  \\
        &DeepLog  & 0.18  & 0.82  & 0.31 \\
        &LogAnomaly & 0.23  & 0.80 & 0.36 \\
        GAIA&PLELog & 0.81 & 0.86 & 0.84\\
        &LogRobust  & 0.83  & 0.94  & 0.88 \\
        &ChatGPT  & 0.68  & 0.75  & 0.71 \\
        &\ourmethod{}  & \textbf{0.89}  & \textbf{0.98} & \textbf{0.93} \\
  
        \bottomrule
    \end{tabular}}
    \caption{Results of different methods on GAIA.}
    \label{GAIA}
    \end{center}
\end{table}
In Table~\ref{GAIA}, \ourmethod{} still achieves the best performance among these baselines. Besides, \ourmethod{} is stably running over 3000 hours on this system, which further demonstrates the stability of the model.

%% file: Conclusion.tex
\section{Conclusions}


In this paper, we propose \ourmethod{}, a pre-train and tuning pipeline for log anomaly detection, which contains the pre-training stage and the adapter-based tuning stage. Besides, the Log-Attention module is proposed  to better encode the information of parameters. Extensive experiments show that our \ourmethod{}, with fewer trainable parameters and lower training costs, outperforms all previous baselines,
which demonstrates the effectiveness of our \ourmethod{}.

%% file: Acknowledgement.tex
\section{Acknowledgments}
This work was supported in part by the National Natural Science Foundation of China (Grant Nos. 62276017, U1636211, 61672081), and the Fund of the State Key Laboratory of Software Development Environment (Grant No. SKLSDE-2021ZX-18).